\documentclass[letterpaper]{article}
\usepackage{uai2018}
\usepackage[margin=1in]{geometry}

\usepackage{times}
\usepackage{microtype}
\usepackage{graphicx}
\usepackage{subfigure}
\usepackage{booktabs} % for professional tables
\usepackage{natbib} 
\usepackage{amsmath}
\usepackage{amssymb}
\usepackage[ansinew]{inputenc}
\usepackage{bm}
\usepackage{mathtools}
\usepackage{hyperref}
\usepackage{natbib}
\usepackage{tikz}
\usetikzlibrary{arrows,shapes,shadows,calc,decorations.pathreplacing,positioning}
\usepackage{pgfplots}
\pgfplotsset{compat=newest}
\pgfplotsset{plot coordinates/math parser=false}
\newlength\figureheight
\newlength\figurewidth 
\newlength\globalheight

\newcommand{\boldalpha}{\bm{\alpha}}
\newcommand{\boldA}{\mathbf{A}}
\newcommand{\boldK}{\mathbf{K}}
\newcommand{\boldf}{\mathbf{f}}
\newcommand{\eye}{\mathbf{I}}
\newcommand{\boldy}{\mathbf{y}}
\newcommand{\boldX}{\mathbf{X}}
\newcommand{\boldSigma}{\bm{\Sigma}}
\newcommand{\boldPhi}{\bm{\Phi}}

\newcommand{\boldz}{\mathbf{z}}
\newcommand{\boldx}{\mathbf{x}} % input space of the outputs
\newcommand{\dif}{\mathrm{d}}   % differential symbol

\pdfoutput=1

\title{Fast Kernel Approximations for Latent Force Models and Convolved Multiple-Output Gaussian processes}

\author{} % LEAVE BLANK FOR ORIGINAL SUBMISSION.
\author{ {\bf Cristian Guarnizo} \\
Faculty of Engineering\\
Universidad Tecnol\'ogica de Pereira\\
Pereira, Colombia, 660003 \\
\And
{\bf Mauricio A. \'Alvarez}\\
Department of Computer Science\\
The University of Sheffield \\
Sheffield, UK, S1 4DP
}

\begin{document}

\maketitle

\begin{abstract}
A latent force model is a Gaussian process with a covariance function inspired by a differential operator. Such covariance function is obtained by performing convolution integrals between Green's functions associated to the differential operators, and covariance functions associated to latent functions. In the classical formulation of latent force models, the covariance functions are obtained analytically by solving a double integral, leading to expressions that involve numerical solutions of different types of error functions. In consequence, the covariance matrix calculation is considerably expensive, because it requires the evaluation of one or more of these error functions. In this paper, we use random Fourier features to approximate the solution of these double integrals obtaining simpler analytical expressions for such covariance functions. We show experimental results using ordinary differential operators and provide an extension to build general kernel functions for convolved multiple output Gaussian processes.
\end{abstract}

\section{INTRODUCTION}
%\section{Introduction}
Latent force models (LFMs) \citep{Alvarez:lfm09} are a type of multiple-output Gaussian processes (GPs) where the covariance
function has been derived from physical models. In particular, LFMs assume that each output $\{f_d(t)\}_{d=1}^D$ can be
expressed as the convolution integral of a latent function $u(t)$, and a Green's function $G_d(t)$ associated to a
linear dynamical system, one per output, $f_d(t) = \int_{0}^t G_d(t-\tau)u(\tau)d\tau$.  Such representation for
$f_d(t)$ introduces a dependency between outputs $f_d(t)$ and $f_{d'}(t)$. For example, if we assume that $u(t)$ follows
a Gaussian process prior with zero mean function and covariance $k(t,t')$, due to the linearity of the integral
transform, $f_d(t)$ and $f_{d'}(t)$ are jointly Gaussian with a cross-covariance function given as
$k_{f_d, f_{d'}} (t,t') = \int_{0}^{t} G_d(t-\tau)\int_{0}^{t'}G_{d'}(t'-\tau') k(\tau,\tau') \dif{\tau'}\dif{\tau}$.

LFMs have been used for uncovering the dynamics of transcription factors in a gene network \citep{Gao:sim:2008}, for
extrapolating human motion from motion capture data \citep{Alvarez:lfm:pami:2013}, for segmenting motor primitives in
humanoid robotics \citep{Alvarez:switched11}, for modeling the thermal properties of buildings
\citep{Ghosh:LFMBuildings:2015}, among several other applications for which prior knowledge of a mechanistic model can
be coded in the covariance function of a GP. By including physics in the covariance function of a GP, we grant
extrapolation abilities to an otherwise interpolation only-model.

In a classical latent force model, the covariance of the latent function $k(t,t')$ follows an Exponentiated Quadratic
(EQ) form, leading to analytical solutions for the cross-covariances $k_{f_d, f_{d'}} (t,t')$. However, these solutions
are computationally expensive since they involve calculating functions that can only be obtained by numerical
methods. For example, using the second order LFM introduced in \citet{Alvarez:lfm09}, involves computing the error
function $\operatorname{erf}(\cdot)$ with a complex argument or the Faddeeva function, that require the
evaluation of numerical integrals that are expensive to compute.

In this work, we use random Fourier features (RFF) \citep{Rahimi:KFF:2007} to reduce the mathematical complexity of the
expressions involved in the covariance functions of the LFM. In particular, we approximate the calculation of the EQ
kernel, with a representation that involves its probability density via the Bochner's theorem. Such
representation for the covariance of $k(\tau,\tau')$ transforms the double integral for $k_{f_d, f_{d'}} (t,t')$ into
two separate integrals that can easily be solved using the Laplace or Fourier transforms. Once the inner integrals are solved (the
integrals that depend on $\tau$ and $\tau'$), the remaining integral is solved using a Monte
Carlo approximation with $S$ samples. The quality of the approximation of the cross-covariances $k_{f_d, f_{d'}} (t,t')$
will depend, then, on the number of samples $S$ used. Additionally, by representing the latent force model kernel using
a sum of basis functions, we are able to reduce the computational complexity of inverting the $ND\times ND$
kernel matrix obtained from the multiple outputs, assuming that each output has $N$ data observations.

Following a similar procedure, we also introduce a random Fourier feature approximation for the more general convolved
multiple output Gaussian process kernel, a model that can be used for multiple-output with no parti\-cular known
dynamics. 

% Furthermore, we can couple that new expressions for fast kernel
% building in LFMs with schemes for computational efficiency in multi-output Gaussian processes \citep{Alvarez:inducing10}
% to speed up even more the inference process.
\section{LATENT FORCE MODELS}
%\section{Latent force models}
% In this section, we provide a short description of latent force models \citet{Alvarez:lfm:pami:2013}, and convolved
% Gaussian process models for multiple outputs. We then follow \citet{Yang:2015} for a brief description on random Fourier features.

% \subsection{Latent force models}
Latent force models are Gaussian processes for multiple outputs with the characteristic that their covariance function involves ordinary or partial differential equations. In particular, LFMs assume that each output $\{f_d(t)\}_{d=1}^D$ can be described using
\begin{align*}
\mathcal{D}_d\{f_d(t)\} = u(t),
\end{align*}
where $\mathcal{D}_d$ is the differential operator associated to a linear ordinary differential equation (ODE) or a
linear partial differential equation (PDE), and $u(t)$ is the excitation function. LFMs assume that $u(t)$ is unknown and place a Gaussian process prior over it. 
The solution for $f_d(t)$ follows as
\begin{align}
f_d(t) & = \int_{0}^t G_d(t-\tau)u(\tau)d\tau,\label{basicLFMQ1eq}
\end{align}
where $G_d(\cdot)$ corresponds to the Green's function associated to the differential operator $\mathcal{D}_d$. The
latent force or function $u(t)$ is unobserved, and follows a Gaussian process prior with zero mean function, and
covariance function given by $k(t, t')$. Since $u(t)$ is being transformed by a linear operator, $f_d(t)$ also follows a
Gaussian process with covariance function $k_{f_d,f_{d}}(t,t')$. Furthermore, since all $f_d(t)$ have a common input
$u(t)$, it is also possible to compute a cross-covariance function between $f_d(t)$, and $f_{d'}(t')$,
$k_{f_d, f_{d'}}(t,t')$.

Equation \eqref{basicLFMQ1eq} can be extended to include additional latent functions with different characteristics, leading to express each output as 
\begin{align*}
f_d(t) & = \sum_{q=1}^Q S_{d,q}\int_{0}^t G_d(t-\tau)u_q(\tau)d\tau,%\label{basicLFMQgt1eq}
\end{align*}
where there are $Q$ latent functions or forces $\{u_q(t)\}_{q=1}^Q$, and $S_{d,q}$ is a sensitivity parameter that
accounts for the influence of force $u_q(t)$ over output $d$. Assuming the independence of these latent
forces and that they all follow Gaussian process priors with covariance functions $k_q(t,t')$, it is possible to compute
the cross-covariance functions $k_{f_d, f_{d'}}(t,t')$, $\forall\,d,d'=1\ldots,D$. The following general expression can
be used to build the covariance $k_{f_d, f_{d'}}(t,t')$ of a LFM
\begin{align}\label{eq:kff}
\sum_{q=1}^{Q} S_{d,q}S_{d',q}& \int_{0}^{t} G_d(t-\tau)\int_{0}^{t'}G_{d'}(t'-\tau')\times \nonumber\\ 
&k_{q}(\tau,\tau') \dif{\tau'}\dif{\tau}.
\end{align}
Depending on the form for the covariance function for $k_{q}(t,t')$, it is possible to find a closed-form expression for
$k_{f_d,f_{d'}}(t,t')$.  A common option for $k_{q}(\tau,\tau')$ is the Exponentiated Quadratic form
\begin{align*}
  k_q(\tau, \tau') = \exp\left(-\frac{(\tau-\tau')^2}{\ell_q^2}\right),
\end{align*}
where $\ell_q$ is known as the length-scale parameter.

LFMs have mostly being used for multiple output regression. In this case, the observed output $d$, $y_d(t)$, is assumed to follow a
Gaussian likelihood, $y_d(t) = f_d(t) + \epsilon_d,$ where $\epsilon_d\sim\mathcal{N}(0, \sigma_d^2)$.
%\begin{align*}
%k_{q}(\tau,\tau') = 
%\end{align*}
%e^{-\frac{(\tau-\tau')^2}{\ell_q^2}} = \int p(\lambda) e^{j(\tau-\tau')\lambda} d\lambda,
% \paragraph{Second-order ODE}
%\subsection{Convolved multiple output Gaussian processes}
%\subsection{Feature expansions for kernel functions}

\section{FEATURE EXPANSIONS FOR KERNELS DERIVED FROM LATENT FORCE MODELS}
%\section{Feature expansions for kernels derived from latent force models}\label{rfflfm}
\label{rfflfm}
In order to scale kernel machines, \citet{Rahimi:KFF:2007} introduced the idea of random Fourier features to approximate
a kernel function using inner products between basis functions. Parameters of these basis functions are sampled from a
distribution associated to the kernel function. We are particularly interested in the approximation for the EQ kernel,
which has been commonly used in LFMs. The idea is to replace the EQ kernel that is usually assumed for
$k_{q}(\tau,\tau') $ by providing a random Fourier feature representation for it via the Bochner's theorem,
\begin{align}
k_{q}(\tau,\tau') = e^{-\frac{(\tau-\tau')^2}{\ell_q^2}} = \int p(\lambda) e^{j(\tau-\tau')\lambda} d\lambda, \label{eq:Bochner}
\end{align}
where $p(\lambda) = \mathcal{N}(\lambda|0, \frac{2}{\ell_q^2})$. A key insight from \citet{Rahimi:KFF:2007} was to use a
finite approximation for $k_{q}(\tau,\tau') $ by using Monte Carlo sampling to solve the above integral over $\lambda$,
\begin{align*}
  k_{q}(\tau,\tau') & \approx \frac{1}{S}\sum_{s=1}^S e^{j\lambda_s\tau}e^{-j\lambda_s\tau'}, \\
  &= \frac{1}{S}\sum_{s=1}^Sv(\tau, \lambda_s)v^*(\tau, \lambda_s),
\end{align*}
where $S$ is the number of Monte Carlo samples, $v(\tau,\lambda_s)$ is a basis function with parameter $\lambda_s$,
$v^*(\tau, \lambda_s)$ is the complex conjugate of $v(\tau,\lambda_s)$, and $\lambda_s\sim p(\lambda)$. Since the kernel function is a real function,
the real part of the product $v(\tau, \lambda_s)v^*(\tau, \lambda_s)$ is used instead. 

Using the expression for $k_{q}(\tau,\tau')$ in Eq. \eqref{eq:Bochner} inside the expression for the 
cross-covariance function for the LFM, $k_{f_d f_{d'}}(t,t')$, we get
\begin{align*}
 \sum_{q=1}^{Q} S_{d,q}S_{d',q} & \int_{0}^{t} G_d(t-\tau)\int_{0}^{t'}G_{d'}(t'-\tau') \times\\
 & \int p(\lambda) e^{j(\tau-\tau')\lambda} d\lambda\dif{\tau'}\dif{\tau}.
\end{align*}
Organizing the above expression we obtain
\begin{align}\label{eq:vvt}
\sum_{q=1}^{Q} S_{d,q}S_{d',q}\int p(\lambda) v_d(t, \theta_d \lambda)
                       v_{d'}^{*}(t',\theta_{d'},\lambda) d\lambda,
\end{align}
with
\begin{align*}
v_d(t,\theta_d, \lambda)& = \int_{0}^{t} G_d(t-\tau)e^{j\lambda\tau} \dif{\tau}, 
\end{align*}
where $\theta_d$ makes reference to the parameters of the Green's function $G_d(\cdot)$. Also,
$v_{d'}^{*}(t',\theta_{d'},\lambda)$ is the complex conjugate for $v_{d'}(t',\theta_{d'},\lambda)$. 
The integrals over $t$ and $t'$ above can be solved using the Laplace transform $\mathcal{L}\{\cdot\}$
\begin{align*}
v_d(t, \theta_d, \lambda)& = \mathcal{L}^{-1}\mathcal{L}\bigg\{\int_{0}^{t} G_d(t-\tau)e^{j\lambda\tau} \dif{\tau}\bigg\}\\& = 
\mathcal{L}^{-1}\bigg\{\mathcal{G}_d(s)\mathcal{L}\big\{e^{j\lambda\tau}\big\}\bigg\},
\end{align*}
where $\mathcal{G}_d(s)$ is the Laplace transform for $G_d(t)$. The operator $\mathcal{L}^{-1}\{\cdot\}$ refers to the inverse
Laplace transform. Furthermore, notice that when $G_{d'}(\cdot)$ is a real function, we can compute
$v^{*}_{d'}(t', \theta_{d'}, \lambda)= v_{d'}(t', \theta_{d'}, -\lambda)$. %CGL: I think this is always valid, or are there any G(t) that is non-real?

Similarly to \citet{Rahimi:KFF:2007}, we use Monte Carlo sampling to approximate the integral over $\lambda$ in
Eq. \eqref{eq:vvt}, leading to
\begin{align*}
\sum_{q=1}^{Q} \frac{S_{d,q}S_{d',q}}{S} \left[\sum_{s=1}^S v_d(t, \theta_d, \lambda_s) v^{*}_{d'}(t', \theta_{d'}, \lambda_s)\right],
\end{align*}  
where $\lambda_s\sim p(\lambda)$.

The steps to compute a RFF approximation of the LFM kernel are
\begin{enumerate}
\item Compute $v_d(t, \theta_d, \lambda) = \int_{0}^{t} G_d(t-\tau)e^{j\lambda\tau} \dif{\tau}$ using the Laplace transform.
\item Compute the RFF approximation for the LFM covariance function $k_{f_d f_{d'}}(t,t')$ using
\begin{align*}
  \sum_{q=1}^{Q} \frac{S_{d,q}S_{d',q}}{S} \left[\sum_{s=1}^S v_d(t, \theta_d, \lambda_s) v^{*}_{d'}(t', \theta_{d'}, \lambda_s)\right],
\end{align*}
where $\lambda_s\sim p(\lambda)$. The distribution we use to sample from, $p(\lambda)$, depends on the kernel assumed for the latent forces $u_q(t)$.
\end{enumerate}
Interestingly, $v_{d}(t, \theta_{d}, \lambda)$ represents the response of the dynamical system to the excitation
$e^{j\lambda t}$ up to time $t$. We will occasionally refer to this random feature as a \textit{random Fourier response feature} (RFRF).

In different applications of LFMs, we need to perform inference over the latent forces $u_{q}(t)$. Inference over $u_{q}(t)$ requires the evaluation of the cross-covariance functions $k_{f_d, u_q}(t,t')$. Such cross-covariances are also important in schemes that reduce computational complexity in convolved multiple output Gaussian processes, where the underlying process $u_{q}(t)$ evaluated at a discrete set of input locations serve the purpose of \textit{inducing variables} \citep{Alvarez:inducing10, Alvarez:CompEfficient:jmlr:2011}. The approximation of $k_{f_d,u_q}(t,t')$ using RFFs is given by
\begin{align*} k_{f_d, u_q}(t,t') 
%&= \frac{1}{S} \sum_{s=1}^{S} \int_{0}^{t} G_d(t-\tau) e^{j(\tau-t')\lambda} \dif \tau\\
&= \frac{1}{S} \sum_{s=1}^{S} v_d(t, \theta_d, \lambda_s) e^{-j\lambda_s t'}.
\end{align*}

\section{HYPERPARAMETER SELECTION AND COMPUTATIONAL COMPLEXITY}
\label{sec:hyper}
Let us assume, we are given observations $\{ \boldy, \boldX \} = \{ \boldy_d, \boldX_d \}_{d=1}^{D}$ ( each
$\boldy_d \in \mathbb{R}^{N}$ and $\boldX_d \in \mathbb{R}^{N\times p}$), and we want to learn the hyperparameters of
the kernel function, $\{\{\theta_d,\sigma_d^2\}_{d=1}^D,\{\ell_q\}_{q=1}^Q\}$, that allow us to explain $\boldy$. With
that in mind, the hyperparamters can be learned from the log-marginal likelihood \citep{Rasmussen:book06}
\begin{align}
\log p(\boldy|\boldX) =& - \frac{ND}{2}\log(2\pi)-\frac{1}{2}  \boldy^{\top}(\boldK_{\boldf, \boldf}+\boldSigma)^{-1} \boldy\notag\\
                       &-\frac{1}{2}\log\left| \boldK_{\boldf, \boldf} + \boldSigma \right|,\label{eq:full:marginal}
\end{align}
where $\boldSigma$ is a diagonal matrix containing the variances of the noise level per output, and
$\boldK_{\boldf,\boldf}\in \mathbb{R}^{ND\times ND}$ is a block-wise matrix with blocks calculated using
\eqref{eq:kff}. As it is usual, we can use a gradient-based optimization procedure to estimate the hyperparameters that
maximize the log-marginal likelihood leading to the infamous computational complexity of $\mathcal{O}(D^3N^3)$. 

However, notice that by the elegance of the RFF representation, the covariance matrix can instead be approximated as
$\boldK_{\boldf,\boldf} = \mathbb{R}\left\{\boldPhi \boldPhi^{\mathsf{H}}\right\}$, where
$\boldPhi \in \mathbb{C}^{ND\times QS}$ has entries $v_d(t,\theta_d, \lambda_s)$, and
$\boldPhi^{\mathsf{H}} $ is the conjugate transpose of $\boldPhi$. Furthermore, the
covariance matrix can be re-written as $\boldK_{\boldf,\boldf} = \bm{\Phi}_c\bm{\Phi}_c^{\top}$, with
$ \bm{\Phi}_c = [\mathbb{R}\{\bm{\Phi}\}\;\;\mathbb{I}\{\bm{\Phi}\}]\in \mathbb{C}^{ND\times 2QS}$. Using the matrix inversion and determinant
lemmas, we express the log-marginal likelihood as
\begin{align}
  \log p(\boldy|\boldX) = & -\frac{1}{2} \log \left| \boldSigma \right| - \frac{1}{2}\left( \boldy^{\top}\boldSigma^{-1}
                            \boldy - \boldalpha^{\top} \boldA^{-1} \boldalpha \right)\notag \\
                          & - \frac{1}{2}\log\left|\boldA\right| - \frac{ND}{2} \log(2\pi),\label{eq:low:rank:marginal}
\end{align}
with $\boldA = \eye+\boldPhi_c^{\top}\boldSigma^{-1}\boldPhi_c$ and
$\boldalpha = \boldPhi_c^{\top} \boldSigma^{-1} \boldy$, effectively reducing computational complexity from
$\mathcal{O}(D^3N^3)$ to $\mathcal{O}(DN Q^2S^2)$, which is now linear with respect to the data size.

Alternatively, one could couple the computation of the kernel functions $k_{f_d,f_{d'}}(t,t')$ and $k_{f_d,u_{q}}(t,t')$
through random Fourier response features, with \textit{(i)} any of the different computationally efficient approximations for
optimizing the log-marginal likelihood in convolved
multiple-output Gaussian process \citep{Alvarez:CompEfficient:jmlr:2011}, or \textit{(ii)} a lower bound on the
log-marginal likelihood through a variational approximation \citep{Alvarez:inducing10}. Both styles of approximations
require the specification of $K$ inducing variables. 

%\paragraph{Variational inducing variables}
%
%\begin{align*} \log p(\boldy|\boldX) = & -\frac{1}{2} \left| \boldSigma \right| - \frac{1}{2}\left( \boldy^{\top}\boldSigma^{-1} \boldy - \tilde{\boldalpha}^{\top} \tilde{\boldA}^{-1} \tilde{\boldalpha} \right) \\ &+ \frac{N}{2} \log(2\pi) - \frac{1}{2}\log\left|\tilde{\boldA}\right|
%\end{align*}

\section{FAST KERNEL BUILDING FROM ORDINARY DIFFERENTIAL EQUATIONS}
%\section{Fast kernel building from ODEs}
Let us assume we are interested in analyzing an ODE of order $P$ given as
\begin{align*}
\mathcal{D}^{(P)}_d\{f_d(t)\} = \sum_{q=1}^Q S_{d,q}u_q(t),
\end{align*}
where the differential operator $\mathcal{D}^{(P)}_d$ is defined as
\[\mathcal{D}^{(P)}_d = a_0 \frac{\dif^{P}}{\dif t^{P}} + a_1 \frac{\dif^{P-1}}{\dif t^{P-1}} + \ldots + a_{P-1}\frac{\dif}{\dif t} + a_P.
\]
The Laplace transform of the Green's function $G_d(t)$ for the above ODE can be found as
\begin{align}\label{eq:ode:roots}
  \mathcal{G}_d(s) &= \frac{1}{a_0} \frac{1}{s^P+\frac{a_1}{a_0}s^{P-1}+\ldots + \frac{a_P}{a_0}}\\
  &  = \frac{1}{a_0} \frac{1}{(s-s_1)(s-s_2)\ldots(s-s_P)}, \nonumber
\end{align}
where the $s_i$'s represent the roots of the polynomial given in the denominator of \eqref{eq:ode:roots}. Additionally,
the Laplace transform for $\mathcal{L}\{e^{j\lambda\tau}\} = \frac{1}{s-j\lambda}$. We can use a partial-fraction
expansion for $\mathcal{G}_d(s)$, and then apply the inverse Laplace transform over the product
$\mathcal{G}_d(s) \mathcal{L}\{e^{j\lambda\tau}\}$ to find $v_d(t,\theta_d, \lambda)$.

Interestingly, if all the roots $s_1,\ldots,s_P$ are distinct real or distinct complex, and $s_{P+1} =
j\lambda$ (the additional root obtained from $\mathcal{L}\{e^{j\lambda\tau}\}$), the random Fourier response feature
$v_d(t,\theta_d, \lambda)$  can be expressed as 
\begin{align}
\frac{1}{a_0} \mathcal{L}^{-1}\left\{ \sum_{p=1}^{P+1} \frac{A_p}{(s-s_p)}
\right\} \nonumber = \frac{1}{a_0} \sum_{p=1}^{P+1} A_pe^{s_pt},
\nonumber
\end{align}
where each coefficient $A_p$ is calculated as
\begin{equation}\label{eq:Ap} A_p = \frac{1}{\prod_{\forall i \neq p}(s_p-s_i)},
\end{equation}
and, as before, $s_{P+1}=j\lambda$.

Next, we show some examples of the expressions obtained for the random Fourier response features associated to the ODE of first and second orders.  Besides, for all ODE experiments the hyperparameters are learned using the variational approach described in \citet{Alvarez:inducing10} and they were carried out using a single core of an AMD FX-8350 @ 4.0 GHz. We also include measures of the time required to evaluate the objective function and its gradients to compare the time cost induced by the evaluation of the different covariance functions. Code to replicate the following experiments is available at \url{github.com/cdguarnizo/kff_lfm}.
\subsection{FIRST-ORDER MODEL (ODE1)}
For the first-order ODE we have the following equation
\begin{align*}
\mathcal{D}^{(1)}_d\{f_d(t)\} = \frac{\dif f_d(t)}{\dif t} + \gamma_df_d(t) = \sum_{q=1}^{Q} S_{d,q}u_q(t),
\end{align*}
%and its Green's function is defined as $G_d(t) = e^{-\gamma_dt}$, 
from which the Laplace transform is given by
$\mathcal{G}_d(s) = \frac{1}{s+\gamma_d} $. We then have $s_1 = -\gamma_d$, and $s_2=j\lambda$.
The random Fourier response feature for the $d$-th output function of a first-order ODE is obtained as
%\begin{align*}
%&v_d(t, \theta_d, \lambda) \\ & = 
%\mathcal{L}^{-1}\bigg\{\mathcal{G}_d(s)\mathcal{L}\big\{e^{j\lambda\tau}\big\}\bigg\}\\& = 
%\mathcal{L}^{-1}\bigg\{\frac{1}{s+\gamma_d} \frac{1}{s-j\lambda}\bigg\}
%\end{align*}
\begin{align*}
  v_d^{(1)}(t,\theta_d,\lambda) &= A_1 e^{s_1 t} + A_2 e^{s_2 t} \\
  &= -\frac{e^{-\gamma_d t}}{\gamma_d+j\lambda} +
  \frac{e^{j\lambda t}}{\gamma_d+j\lambda}\\
  &=\frac{e^{j\lambda t}-e^{-\gamma_d t}}{\gamma_d+j\lambda}.
\end{align*}
Next, we compare the performance of the first order ODE described in \citet{Gao:sim:2008} with the kernel obtained by using the above random Fourier response feature for interpolation of Air temperature.

\newcommand{\plotlfm}[1]{\input{figures/weather_out#1_sim.tikz}}
\newcommand{\plotkfff}[1]{\input{figures/weather_out1_kff_sim_S#1.tikz}}
\newcommand{\plotkffs}[1]{\input{figures/weather_out2_kff_sim_S#1.tikz}}
\begin{figure*}[h!]
	\centering
	\setlength{\tabcolsep}{1pt}
	\figurewidth = .59\columnwidth
	\figureheight = .3\columnwidth
	\begin{tabular}{rccc}
		& ODE1 & ODE1 + S100 & ODE1 + S10 \\
		\raisebox{.6cm}{\rotatebox{90}{Cambermet}} & % This file was created by matlab2tikz.
%
\begin{tikzpicture}
\tikzset{font=\small}
\begin{axis}[%
width=\figurewidth,
height=\figureheight,
at={(0\figurewidth,0\figureheight)},
scale only axis,
axis on top,
xmin=0,
xmax=3,
xtick={0,1,2,3},
xticklabels={},
ymin=-4,
ymax=3,
ytick style={draw=none},
ytick={-4,-2,0,2},
axis background/.style={fill=white}
]
\addplot [forget plot] graphics [xmin=-0.04, xmax=3.04, ymin=-4.05, ymax=3.05] {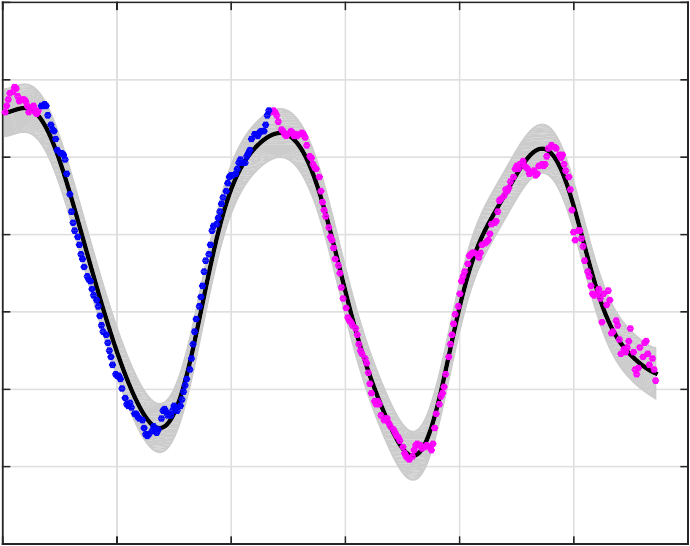};
\end{axis}
\end{tikzpicture}%
 & % This file was created by matlab2tikz.
%
\begin{tikzpicture}
\tikzset{font=\small}
\begin{axis}[%
width=\figurewidth,
height=\figureheight,
at={(0\figurewidth,0\figureheight)},
scale only axis,
axis on top,
xmin=0,
xmax=3,
xtick={0,1,2,3},
xticklabels={},
ymin=-4,
ymax=3,
ytick style={draw=none},
ytick={-4,-3,-2,-1,0,1,2,3},
yticklabels={},
axis background/.style={fill=white}
]
\addplot [forget plot] graphics [xmin=-0.04, xmax=3.04, ymin=-4.05, ymax=3.05] {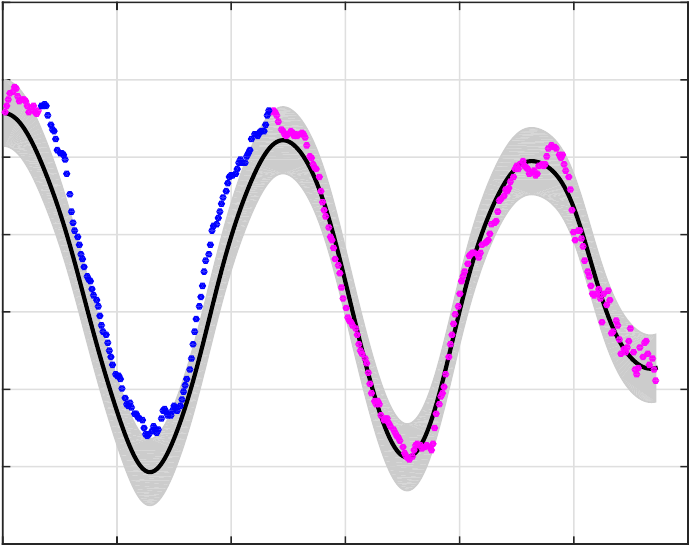};
\end{axis}
\end{tikzpicture}%
 & % This file was created by matlab2tikz.
%
\begin{tikzpicture}
\tikzset{font=\small}
\begin{axis}[%
width=\figurewidth,
height=\figureheight,
at={(0\figurewidth,0\figureheight)},
scale only axis,
axis on top,
xmin=0,
xmax=3,
xtick={0,1,2,3},
xticklabels={},
ymin=-4,
ymax=3,
ytick style={draw=none},
ytick={-4,-3,-2,-1,0,1,2,3},
yticklabels={},
axis background/.style={fill=white}
]
\addplot [forget plot] graphics [xmin=-0.04, xmax=3.04, ymin=-4.05, ymax=3.05] {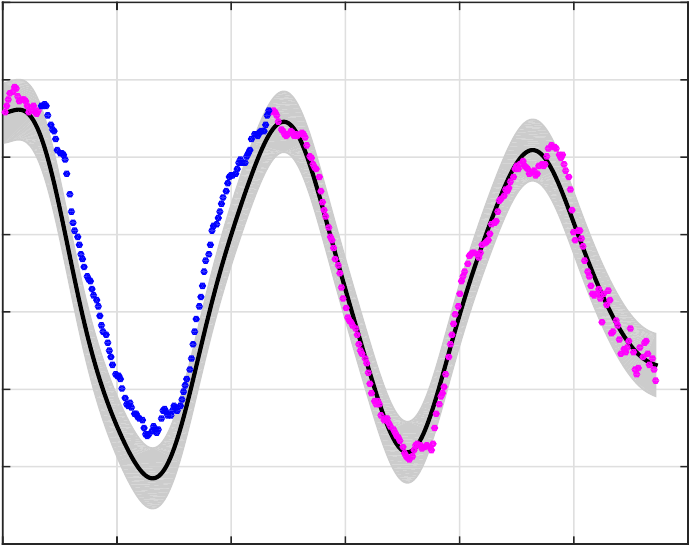};
\end{axis}
\end{tikzpicture}%
 \\
		\raisebox{1cm}{\rotatebox{90}{Chimet}} & % This file was created by matlab2tikz.
%
\begin{tikzpicture}
\tikzset{font=\small}
\begin{axis}[%
width=\figurewidth,
height=\figureheight,
at={(0\figurewidth,0\figureheight)},
scale only axis,
axis on top,
xmin=0,
xmax=3,
xtick={0,1,2,3},
ymin=0,
ymax=100,
ytick style={draw=none},
ytick={0,20,40,60,80,100},
axis background/.style={fill=white}
]
\addplot [forget plot] graphics [xmin=-0.04, xmax=3.04, ymin=-0.05, ymax=100.05] {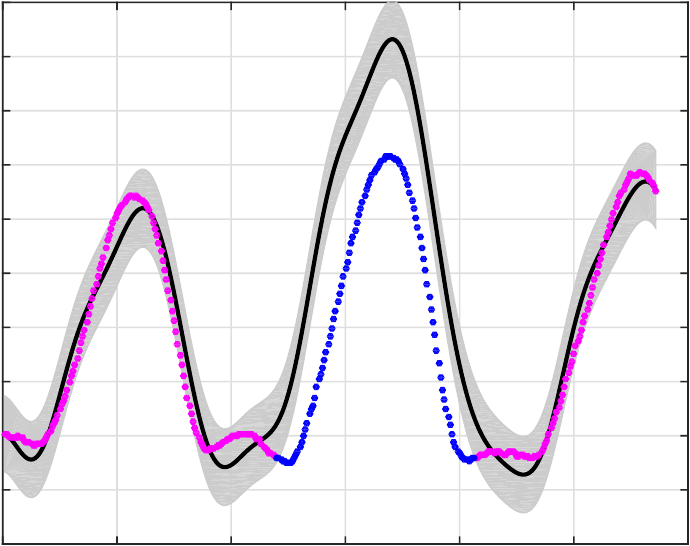};
\end{axis}
\end{tikzpicture}%
  & % This file was created by matlab2tikz.
%
\begin{tikzpicture}
\tikzset{font=\small}
\begin{axis}[%
width=\figurewidth,
height=\figureheight,
at={(0\figurewidth,0\figureheight)},
scale only axis,
axis on top,
xmin=0,
xmax=3,
xtick={0,1,2,3},
ymin=0,
ymax=100,
ytick style={draw=none},
ytick={0,20,40,60,80,100},
yticklabels={},
axis background/.style={fill=white}
]
\addplot [forget plot] graphics [xmin=-0.04, xmax=3.04, ymin=-0.05, ymax=100.05] {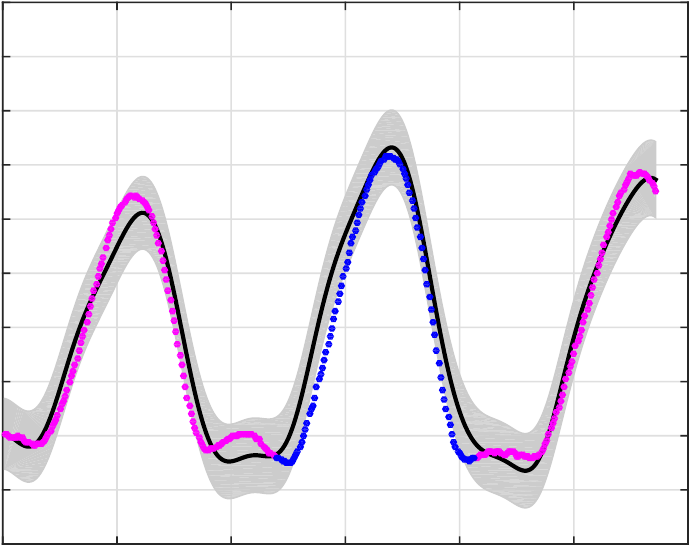};
\end{axis}
\end{tikzpicture}%
 & % This file was created by matlab2tikz.
%
\begin{tikzpicture}
\tikzset{font=\small}
\begin{axis}[%
width=\figurewidth,
height=\figureheight,
at={(0\figurewidth,0\figureheight)},
scale only axis,
axis on top,
xmin=0,
xmax=3,
xtick={0,1,2,3},
ymin=0,
ymax=100,
ytick style={draw=none},
ytick={0,20,40,60,80,100},
yticklabels={},
axis background/.style={fill=white}
]
\addplot [forget plot] graphics [xmin=-0.04, xmax=3.04, ymin=-0.05, ymax=100.05] {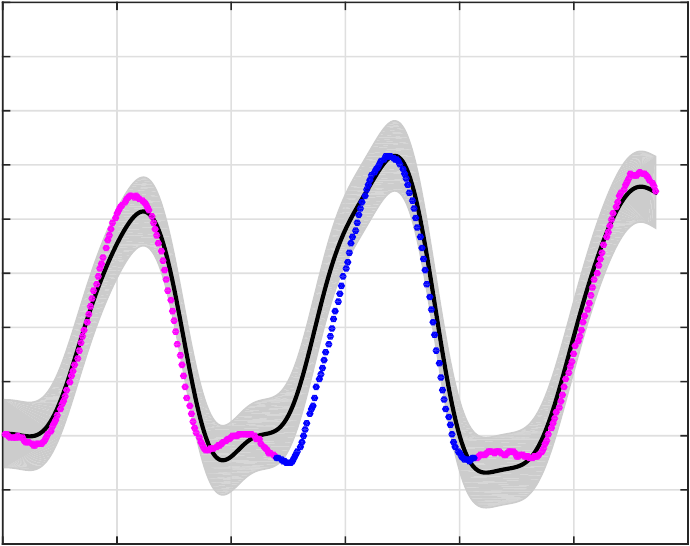};
\end{axis}
\end{tikzpicture}%
 \\  		
		& Time (days) & Time (days) & Time (days)
	\end{tabular}
	\caption{Comparison of the predictive GPs, for the air temperature experiment, using the standard LFM (first column) and the RFRF approximation for $S=100$ (second column) and $S=10$ samples (third column). Training data is represented using red dots and Test data using blue dots. The black line in the mean over the predictive GP function, and the shaded region denotes two times the standard deviation.}
	\label{fig:weather}
\end{figure*}

\paragraph{Air temperature}\label{sec:results:weather}
Here, we consider the problem of modeling and predicting air temperature time series from a network sensor located at the south coast of England. The dataset consists of temperature measurements at four locations known as Bramblemet, Sotonmet, Cambermet and Chimet.
\footnote{Weather data can be found in \url{http://www.bramblemet.co.uk}.} 
The air temperatures are measured during the period from July 10 to July 15, 2013. Specifically, we adopt the same experiment (train and test data) used in \cite{Nguyen:2014} and described in Tab. \ref{tab:weather}. The variational approach is configured with 200 inducing variables, six latent forces and the maximum number of iterations for the optimization procedure is set to 500.

\begin{table}[ht]
	\caption{Number of training and test data-points considered on the air temperature experiment.}
	\label{tab:weather}
	\centering
	\begin{tabular}{llcc}
		\hline
		\hline
		\# & Name & Training & Test\\
		\hline
		1 & Bramblemet & 1425 & 0\\
		2 & Cambermet & 1268 & 173\\
		3 & Chimet & 1235 & 201\\
		4 & Sotonmet & 1097 & 0\\
		\hline
		\hline
	\end{tabular}
\end{table}

Table \ref{tab:weather:res} reports the predictive performance using the covariance functions build from the LFM and the proposed RFRF. Note that for a low number of samples $S$, the proposed approach presents the worst performance. This is because the more samples we use the better the mean of predictive GP is able to fit the coarse behavior from the observed data, as shown in figure \ref{fig:weather}. Interestingly, the RFRF starts to outperform the standard one, using only 50 or 100 samples with about half of the time required by the original covariance function. 

\begin{table}[h]
	\caption{Results on air temperature data.}
	\label{tab:weather:res}
	\setlength{\tabcolsep}{2pt}
	\centering
	\begin{tabular}{c|cc|cc|c}
		\hline
		\hline
		Kernel & \multicolumn{2}{c|}{Cambermet} & \multicolumn{2}{c|}{Chimet} & Time \\
		& NMSE & NLPD  & NMSE & NLPD & [s]\\
		%\hline
		%GG & 0.13 & 1.32 & 0.27 & 1.17 & 1.81 \\
		%GG10 & 0.30 & 1.70 & 1.40 & 1.91 & 1.72 \\
		%GG20 & 0.71 & 2.15 & 0.46 & 1.51 & 2.17 \\
		%GG50 & 0.16 & 1.78 & 1.35 & 1.94 & 2.56 \\
		%GG100 & 0.09 & 1.03 & 0.18 & 0.88 & 3.26 \\
		\hline
		ODE1+S10 & 0.74 & 3.26 & 0.58 & 1.53 & 1.89 \\
		ODE1+S20 & 0.45 & 1.95 & 0.93 & 1.75 & 2.09 \\
		ODE1+S50 & \bfseries{0.08} & \bfseries{1.10} & 0.21 & 1.08 & 2.68 \\
		ODE1+S100 & 0.12 & 1.18 & \bfseries{0.12} & \bfseries{0.82} & 3.93 \\
		ODE1 & 0.11 & 1.37 & 0.19 & 0.99 & 6.28 \\
		\hline
		\hline
	\end{tabular}
\end{table}

\subsection{SECOND-ORDER MODEL (ODE2)}
As a second example of a random Fourier feature representation of a LFM, we use a second-order ordinary differential
operator $\mathcal{D}^{(2)}_d\{\cdot\}$ that represents, e.g., a mass-spring-damper system. The second-order operator is given as
\begin{align*}
	\mathcal{D}^{(2)}_d = m_d\frac{d^2}{dt^2} + c_d\frac{d}{dt} + b_d,
\end{align*}
where $m_d$, $c_d$ and $b_d$ are the mass, damper and spring constants, respectively. From the above equation, we obtain
the Laplace transform of the Green's function as
\begin{align*}
\mathcal{G}_d(s) = \frac{1}{m_d}\frac{1}{s^2+\frac{c_d}{m_d}s+\frac{b_d}{m_d}}.
\end{align*}
Following the procedure described above, it can be shown that the random Fourier response feature for the $d$-th output is given by
\begin{align*}
v_d^{(2)}&(t,\theta_d,\lambda)=&\frac{1}{m_d} \biggl[ A_1e^{s_1t} + A_2e^{s_2 t} + A_3e^{s_3 t} \biggr],
\end{align*}
where 
\[s_1,s_2 = -\frac{c_d}{2m_d} \pm \sqrt{\frac{c_d^2}{4m_d^2} - \frac{b_d}{m_d}},\] are the roots of the polynomial
obtained from the second-order ODE, and $s_3 = j\lambda$ corresponds to the root induced by the excitation
$e^{j\lambda t}$.  Note that the coefficients $A_1$ and $A_2$ were calculated using \eqref{eq:Ap}. Furthermore, if
$c_d^2 > 4m_db_d$ then the roots $s_1$ and $s_2$ are real, and the model's response is known as ``overdamped''. When
$c_d^2 < 4m_db_d$ the roots are a pair of complex conjugates, and the response is known as ``underdamped''.

\begin{figure}[h!]
	\figurewidth = .9\columnwidth
	\figureheight = .4\columnwidth
	\begin{tabular}{l}
	\hspace{1.3cm} ODE2 \hspace{1.7cm} ODE2+S100\\
	\hspace{-.3cm}% This file was created by matlab2tikz.
%
\begin{tikzpicture}
\tikzset{font=\tiny}
\begin{axis}[%
width=0.45\figurewidth,
height=\figureheight,
at={(0\figurewidth,0\figureheight)},
scale only axis,
point meta min=-5.06736512459666,
point meta max=38.7029815931949,
axis on top,
xmin=0.5,
xmax=200.5,
xtick={50,100,150,200},
xticklabels={{1.5},{3.0},{1.5},{3.0}},
y dir=reverse,
ymin=0.5,
ymax=200.5,
ytick={50,100,150,200},
yticklabels={{1.5},{3.0},{1.5},{3.0}},
axis background/.style={fill=white}
]
\addplot [forget plot] graphics [xmin=0.5, xmax=200.5, ymin=0.5, ymax=200.5] {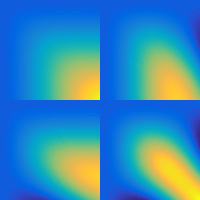};
\end{axis}

\begin{axis}[%
width=0.45\figurewidth,
height=\figureheight,
at={(0.48\figurewidth,0\figureheight)},
scale only axis,
point meta min=-5.40669217617442,
point meta max=36.9400544022306,
axis on top,
xmin=0.5,
xmax=200.5,
xtick={50,100,150,200},
xticklabels={{1.5},{3.0},{1.5},{3.0}},
y dir=reverse,
ymin=0.5,
ymax=200.5,
ytick={50,100,150,200},
yticklabels={},
axis background/.style={fill=white},
colormap={mymap}{[1pt] rgb(0pt)=(0.2081,0.1663,0.5292); rgb(1pt)=(0.211624,0.189781,0.577676); rgb(2pt)=(0.212252,0.213771,0.626971); rgb(3pt)=(0.2081,0.2386,0.677086); rgb(4pt)=(0.195905,0.264457,0.7279); rgb(5pt)=(0.170729,0.291938,0.779248); rgb(6pt)=(0.125271,0.324243,0.830271); rgb(7pt)=(0.0591333,0.359833,0.868333); rgb(8pt)=(0.0116952,0.38751,0.881957); rgb(9pt)=(0.00595714,0.408614,0.882843); rgb(10pt)=(0.0165143,0.4266,0.878633); rgb(11pt)=(0.0328524,0.443043,0.871957); rgb(12pt)=(0.0498143,0.458571,0.864057); rgb(13pt)=(0.0629333,0.47369,0.855438); rgb(14pt)=(0.0722667,0.488667,0.8467); rgb(15pt)=(0.0779429,0.503986,0.838371); rgb(16pt)=(0.0793476,0.520024,0.831181); rgb(17pt)=(0.0749429,0.537543,0.826271); rgb(18pt)=(0.0640571,0.556986,0.823957); rgb(19pt)=(0.0487714,0.577224,0.822829); rgb(20pt)=(0.0343429,0.596581,0.819852); rgb(21pt)=(0.0265,0.6137,0.8135); rgb(22pt)=(0.0238905,0.628662,0.803762); rgb(23pt)=(0.0230905,0.641786,0.791267); rgb(24pt)=(0.0227714,0.653486,0.776757); rgb(25pt)=(0.0266619,0.664195,0.760719); rgb(26pt)=(0.0383714,0.674271,0.743552); rgb(27pt)=(0.0589714,0.683757,0.725386); rgb(28pt)=(0.0843,0.692833,0.706167); rgb(29pt)=(0.113295,0.7015,0.685857); rgb(30pt)=(0.145271,0.709757,0.664629); rgb(31pt)=(0.180133,0.717657,0.642433); rgb(32pt)=(0.217829,0.725043,0.619262); rgb(33pt)=(0.258643,0.731714,0.595429); rgb(34pt)=(0.302171,0.737605,0.571186); rgb(35pt)=(0.348167,0.742433,0.547267); rgb(36pt)=(0.395257,0.7459,0.524443); rgb(37pt)=(0.44201,0.748081,0.503314); rgb(38pt)=(0.487124,0.749062,0.483976); rgb(39pt)=(0.530029,0.749114,0.466114); rgb(40pt)=(0.570857,0.748519,0.44939); rgb(41pt)=(0.609852,0.747314,0.433686); rgb(42pt)=(0.6473,0.7456,0.4188); rgb(43pt)=(0.683419,0.743476,0.404433); rgb(44pt)=(0.71841,0.741133,0.390476); rgb(45pt)=(0.752486,0.7384,0.376814); rgb(46pt)=(0.785843,0.735567,0.363271); rgb(47pt)=(0.818505,0.732733,0.34979); rgb(48pt)=(0.850657,0.7299,0.336029); rgb(49pt)=(0.882433,0.727433,0.3217); rgb(50pt)=(0.913933,0.725786,0.306276); rgb(51pt)=(0.944957,0.726114,0.288643); rgb(52pt)=(0.973895,0.731395,0.266648); rgb(53pt)=(0.993771,0.745457,0.240348); rgb(54pt)=(0.999043,0.765314,0.216414); rgb(55pt)=(0.995533,0.786057,0.196652); rgb(56pt)=(0.988,0.8066,0.179367); rgb(57pt)=(0.978857,0.827143,0.163314); rgb(58pt)=(0.9697,0.848138,0.147452); rgb(59pt)=(0.962586,0.870514,0.1309); rgb(60pt)=(0.958871,0.8949,0.113243); rgb(61pt)=(0.959824,0.921833,0.0948381); rgb(62pt)=(0.9661,0.951443,0.0755333); rgb(63pt)=(0.9763,0.9831,0.0538)},
colorbar right,
colorbar style={anchor=south west, at={(1.03,0)}, height=0.997*\pgfkeysvalueof{/pgfplots/parent axis height}, width = 0.04*\figurewidth }
]
\addplot [forget plot] graphics [xmin=0.5, xmax=200.5, ymin=0.5, ymax=200.5] {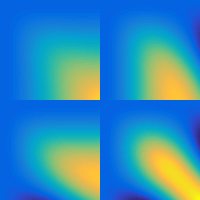};
\end{axis}

\begin{axis}[%
width=1.227\figurewidth,
height=1.227\figureheight,
at={(-0.01\figurewidth,-0.135\figureheight)},
scale only axis,
point meta min=0,
point meta max=1,
xmin=0,
xmax=1,
ymin=0,
ymax=1,
axis line style={draw=none},
ticks=none,
axis x line*=bottom,
axis y line*=left,
clip=false
]
%\node[below, align=left]
%at (rel axis cs:0,0) {$\alpha$};
\node[below right, align=left]
at (rel axis cs:0.43,1.03) {$\mathbf{K}_{\mathbf{f}_1\mathbf{f}_1}$};
\node[below right, align=left]
at (rel axis cs:0.62,1.03) {$\mathbf{K}_{\mathbf{f}_1\mathbf{f}_2}$};
\node[below right, align=left]
at (rel axis cs:0.43,0.05) {$\mathbf{K}_{\mathbf{f}_2\mathbf{f}_1}$};
\node[below right, align=left]
at (rel axis cs:0.62,0.05) {$\mathbf{K}_{\mathbf{f}_2\mathbf{f}_2}$};
\node[below right, align=left]
at (rel axis cs:0.045,1.03) {$\mathbf{K}_{\mathbf{f}_1\mathbf{f}_1}$};
\node[below right, align=left]
at (rel axis cs:0.23,1.03) {$\mathbf{K}_{\mathbf{f}_1\mathbf{f}_2}$};
\node[below right, align=left]
at (rel axis cs:0.045,0.05) {$\mathbf{K}_{\mathbf{f}_2\mathbf{f}_1}$};
\node[below right, align=left]
at (rel axis cs:0.23,0.05) {$\mathbf{K}_{\mathbf{f}_2\mathbf{f}_2}$};
\end{axis}
\end{tikzpicture}%
	\end{tabular}
	\caption[]{Comparison of the covariance matrix evaluation using the standard LFM and the RFRF.}
	\label{fig:Kmatrix}
\end{figure}
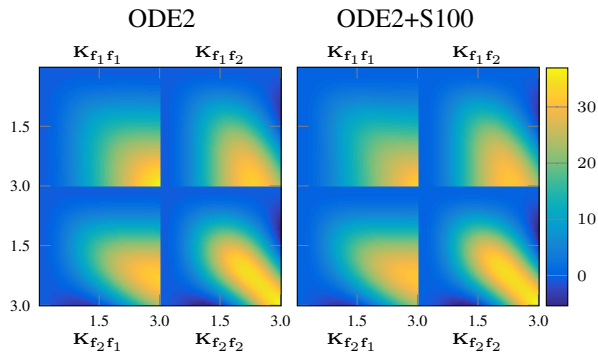

Figure \ref{fig:Kmatrix} shows the covariance matrices for a two-output LFM using the standard expression for the covariance function in \citet{Alvarez:lfm09}, and the kernel obtained by using the random Fourier response features for the ODE2, $v_d^{(2)}(t,\theta_d,\lambda)$, based on $S=100$ samples. In this example, we consider that the first output follows an overdamped response, while the second output has an underdamped response. Additionally, the input times comprises 100 values in the range from 0s to 3s for each output. Just to have a quantitative measure of the approximation obtained by the RFRF approach, the Frobenius norm between the covariance matrices shown in figure \ref{fig:Kmatrix} is 239.1. However, for $S=10^{5}$ samples, the Frobenius norm is 5.8, which states that we are able to reduce the approximation error by the cost of increasing the number of samples.
%\[s_1,s_2 = -\frac{c_d}{2m_d} \pm \sqrt{\frac{c_d^2}{4m_d^2} - \frac{b_d}{m_d}},\]
%are the roots of the polynomial obtained from the second-order ODE and $s_3 = j\lambda$ corresponds to the root induced by the excitation $e^{j\lambda}$. Note that the coefficients $A_i$'s are calculated using \eqref{eq:Ap}. Furthermore, if $c_d^2 > 4m_db_d$ then the roots $s_1$ and $s_2$ are real, and the model's response is known as ``overdamped''. On contrast, when $c_d^2 < 4m_db_d$ the roots are a pair of complex conjugates, and the response is known as ``underdamped''.
%
%We perform a graphical comparison between the covariance matrices of a two output model using the standard LFM and the RFF approximation based on $S=100$ samples, as shown in figure \ref{fig:Kmatrix}. In this example, we consider that the first output follows an overdamped response, while the second output has an underdamped response. Additionally, the time values range from 0s to 3s for both outputs.
%S=100K Frob = 5.8488, S=10K Frob =48.2379
%S=1K Frob = 98.3876, S=100 Frob = 239.1073
Note that the covariance values are similar, indicating that the correlation between the outputs and within each output is preserved and well approximated by the inner products of the random features $v_d^{(2)}(t,\theta_d,\lambda)$.
For the following experiments, we consider two motion capture (MOCAP) datasets,
\footnote{MOCAP datasets are available at \url{http://mocap.cs.cmu.edu/}.} 
which consist of measured joint angles from different types of motions. Additionally, the variational approach is configured with 25 inducing variables, six latent forces and the maximum number of iterations set to 500.

\paragraph{MOCAP - Golf swing}
In this experiment, we consider the movement ``Golf swing'' performed by subject 64 motion 01. From the 62 available channels, we selected 56 each having 448 samples, except for two outputs where 81 consecutive samples were considered for testing purposes. The complete dataset for training consists of 24926 data-points. 
\begin{table}[h!]
	\caption{Results for Golf Swing dataset.}
	\label{tab:mocap:golf}
	\setlength{\tabcolsep}{2pt}
	\centering
	\begin{tabular}{c|cc|cc|c}
		\hline
		\hline
		Kernel & \multicolumn{2}{c|}{root-Ypos} & \multicolumn{2}{c|}{lowerback-Yrot} & Time \\
		& NMSE & NLPD  & NMSE & NLPD & [s]\\
		\hline
%		GG & 0.37 & -2.22 & 9.44 & 19.46 & 0.70 \\
%		\hline
		ODE2+S10 & 0.39 & -2.23 & 0.98 & 2.69 & 2.20 \\
		ODE2+S20 & 0.24 & -2.35 & 1.49 & 4.30 & 3.02 \\
		ODE2+S50 & 0.17 & -2.39 & \bfseries{0.27} & \bfseries{1.17} & 4.59 \\
		ODE2+S100 & 0.12 & \bfseries{-2.45} & 0.32 & 1.34 & 9.31\\
		ODE2 & \bfseries{0.11} & -2.39 & 3.19 & 7.26 & 28.96 \\
		\hline
		\hline
	\end{tabular}
\end{table}

\begin{figure*}[h!]
	\centering
	\setlength{\tabcolsep}{1pt}
	\figurewidth = .59\columnwidth
	\figureheight = .3\columnwidth
	\begin{tabular}{rccc}
		& ODE2 & ODE2+S100 & ODE2+S10\\
		\raisebox{.7cm}{\rotatebox{90}{root-Ypos}} & % This file was created by matlab2tikz.
%
\begin{tikzpicture}
\tikzset{font=\small}
\begin{axis}[%
width=\figurewidth,
height=\figureheight,
at={(0\figurewidth,0\figureheight)},
scale only axis,
axis on top,
xmin=0,
xmax=3,
xtick={0,1,2,3},
xticklabels={},
ymin=-4,
ymax=3,
ytick style={draw=none},
ytick={-4,-2,0,2},
axis background/.style={fill=white}
]
\addplot [forget plot] graphics [xmin=-0.04, xmax=3.04, ymin=-4.05, ymax=3.05] {figures/walk_out1_lfm.pdf};
\end{axis}
\end{tikzpicture}%
 & % This file was created by matlab2tikz.
%
\begin{tikzpicture}
\tikzset{font=\small}
\begin{axis}[%
width=\figurewidth,
height=\figureheight,
at={(0\figurewidth,0\figureheight)},
scale only axis,
axis on top,
xmin=0,
xmax=3,
xtick={0,1,2,3},
xticklabels={},
ymin=-4,
ymax=3,
ytick style={draw=none},
ytick={-4,-3,-2,-1,0,1,2,3},
yticklabels={},
axis background/.style={fill=white}
]
\addplot [forget plot] graphics [xmin=-0.04, xmax=3.04, ymin=-4.05, ymax=3.05] {figures/walk_out1_kfflfm_S100.pdf};
\end{axis}
\end{tikzpicture}%
 & % This file was created by matlab2tikz.
%
\begin{tikzpicture}
\tikzset{font=\small}
\begin{axis}[%
width=\figurewidth,
height=\figureheight,
at={(0\figurewidth,0\figureheight)},
scale only axis,
axis on top,
xmin=0,
xmax=3,
xtick={0,1,2,3},
xticklabels={},
ymin=-4,
ymax=3,
ytick style={draw=none},
ytick={-4,-3,-2,-1,0,1,2,3},
yticklabels={},
axis background/.style={fill=white}
]
\addplot [forget plot] graphics [xmin=-0.04, xmax=3.04, ymin=-4.05, ymax=3.05] {figures/walk_out1_kfflfm_S10.pdf};
\end{axis}
\end{tikzpicture}%
\\
		\raisebox{.5cm}{\rotatebox{90}{lowerback-Yrot}} & % This file was created by matlab2tikz.
%
\begin{tikzpicture}
\tikzset{font=\small}
\begin{axis}[%
width=\figurewidth,
height=\figureheight,
at={(0\figurewidth,0\figureheight)},
scale only axis,
axis on top,
xmin=0,
xmax=3,
xtick={0,1,2,3},
ymin=0,
ymax=100,
ytick style={draw=none},
ytick={0,20,40,60,80,100},
axis background/.style={fill=white}
]
\addplot [forget plot] graphics [xmin=-0.04, xmax=3.04, ymin=-0.05, ymax=100.05] {figures/walk_out2_lfm.pdf};
\end{axis}
\end{tikzpicture}%
 & % This file was created by matlab2tikz.
%
\begin{tikzpicture}
\tikzset{font=\small}
\begin{axis}[%
width=\figurewidth,
height=\figureheight,
at={(0\figurewidth,0\figureheight)},
scale only axis,
axis on top,
xmin=0,
xmax=3,
xtick={0,1,2,3},
ymin=0,
ymax=100,
ytick style={draw=none},
ytick={0,20,40,60,80,100},
yticklabels={},
axis background/.style={fill=white}
]
\addplot [forget plot] graphics [xmin=-0.04, xmax=3.04, ymin=-0.05, ymax=100.05] {figures/walk_out2_kfflfm_S100.pdf};
\end{axis}
\end{tikzpicture}%
 & % This file was created by matlab2tikz.
%
\begin{tikzpicture}
\tikzset{font=\small}
\begin{axis}[%
width=\figurewidth,
height=\figureheight,
at={(0\figurewidth,0\figureheight)},
scale only axis,
axis on top,
xmin=0,
xmax=3,
xtick={0,1,2,3},
ymin=0,
ymax=100,
ytick style={draw=none},
ytick={0,20,40,60,80,100},
yticklabels={},
axis background/.style={fill=white}
]
\addplot [forget plot] graphics [xmin=-0.04, xmax=3.04, ymin=-0.05, ymax=100.05] {figures/walk_out2_kfflfm_S10.pdf};
\end{axis}
\end{tikzpicture}%
\\  		
		& Time (s) & Time (s) & Time (s)
	\end{tabular}
	\caption{Comparison of the predictive GPs, for the Golf swing experiment, using the standard LFM (first column) and the RFRF approximation for $S=100$ (second column) and $S=10$ samples (third column) . Training data is represented using red dots and Test data using blue dots. The black line in the mean over the predictive GP function, and the shaded region denotes two times the standard deviation.}
	\label{fig:golf}
\end{figure*}
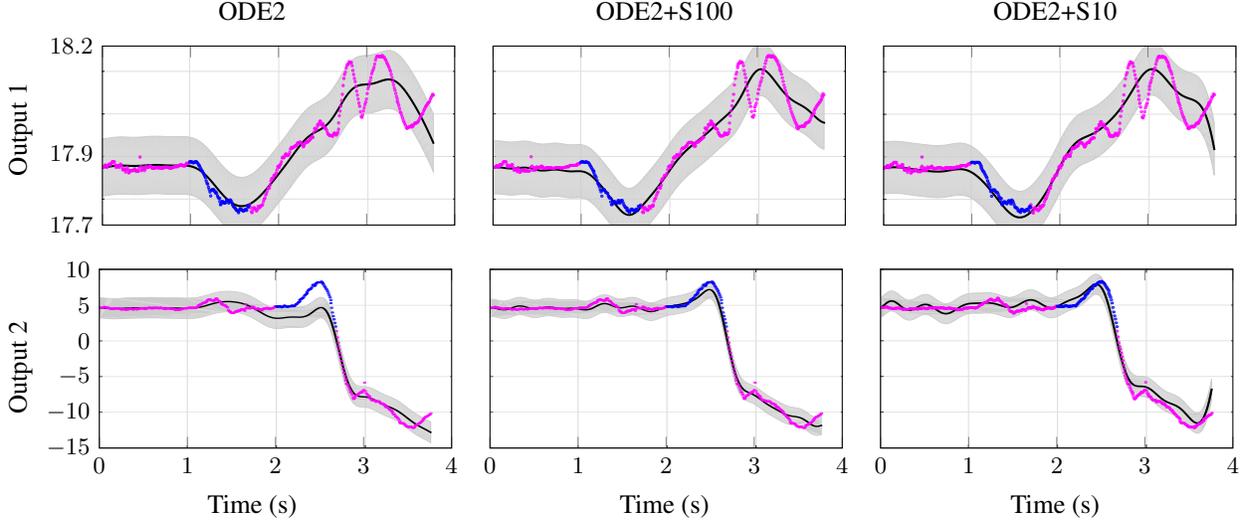

Table \ref{tab:mocap:golf} reports the predictive performance using the covariance functions built from the LFM and the proposed RFRF. In this experiment, the RFRF approximations fit better the testing data for output ``lowerback-Yrot'', as shown in figure \ref{fig:golf}. In contrast, output ``root-Ypos'' testing data is best fitted by the standard LFM. In summary, the models learned using 50 and 100 samples not only performed better than the standard LFM, but also their cost time is reduced by a fraction of three and six, respectively.

\renewcommand{\plotlfm}[1]{\input{figures/Swing_out#1_lfm.tikz}}
\renewcommand{\plotkfff}[1]{\input{figures/Swing_out1_kfflfm_S#1.tikz}}
\renewcommand{\plotkffs}[1]{\input{figures/Swing_out2_kfflfm_S#1.tikz}}

\paragraph{MOCAP - Walk}
For this experiment, we consider the movement ``walk'' from subject 02 motion 01. From the 62 available channels, we selected 48 each having 343 samples, except for 121 and 105 consecutive samples of two outputs that were considered for testing purposes. The complete dataset for training consists of 16238 data-points.
\begin{table}[h!]
	\caption{Results for Walk Dataset.}
	\label{tab:mocap:walk}
	\setlength{\tabcolsep}{2pt}
	\centering
	\begin{tabular}{c|cc|cc|c}
		\hline
		\hline
		Kernel & \multicolumn{2}{c|}{lowerback-Yrot} & \multicolumn{2}{c|}{lradius-Xrot} & Time \\
		& NMSE & NLPD  & NMSE & NLPD & [s]\\
		%\hline
		%GG & 2.79 & 3.25 & 0.03 & 4.61 & 0.54 \\
		\hline
		ODE2+S10 & 0.21 & 5.05 & 0.12 & 1.06 & 1.45 \\
		ODE2+S20 & 0.22 & 2.09 & 0.49 & \bfseries{0.87} & 2.04 \\
		ODE2+S50 & 0.22 & 4.77 & 0.19 & 5.28 & 3.24 \\
		ODE2+S100 & 0.18 & 3.35 & \bfseries{0.09} & 3.86 & 6.09\\
		ODE2 & \bfseries{0.02} & \bfseries{-0.10} & 0.99 & 19.63 & 19.67\\
		\hline
		\hline
	\end{tabular}
\end{table}

Table \ref{tab:mocap:walk} reports the predictive performance for the testing data used in ``walk''  experiment. Output ``lowerback-Yrot'' missing data is best fitted by the standard LFM. However, the testing data for output ``lradius-Xrot'' is best fitted by the proposed RFRF approach, as shown in figure \ref{fig:golf}. Interestingly, for this experiment, the observed data are smooth, which can be fitted with adequate accuracy using 10 or 20 samples using the RFRF approach.

\renewcommand{\plotlfm}[1]{\input{figures/walk_out#1_lfm.tikz}}
\renewcommand{\plotkfff}[1]{\input{figures/walk_out1_kfflfm_S#1.tikz}}
\renewcommand{\plotkffs}[1]{\input{figures/walk_out2_kfflfm_S#1.tikz}}
\begin{figure*}[h!]
	\centering
	\setlength{\tabcolsep}{1pt}
	\figurewidth = .59\columnwidth
	\figureheight = .3\columnwidth
	\begin{tabular}{rccc}
		& ODE2 & ODE2+S100 & ODE2+S10\\
		\raisebox{.3cm}{\rotatebox{90}{lowerback-Yrot}} & % This file was created by matlab2tikz.
%
\begin{tikzpicture}
\tikzset{font=\small}
\begin{axis}[%
width=\figurewidth,
height=\figureheight,
at={(0\figurewidth,0\figureheight)},
scale only axis,
axis on top,
xmin=0,
xmax=3,
xtick={0,1,2,3},
xticklabels={},
ymin=-4,
ymax=3,
ytick style={draw=none},
ytick={-4,-2,0,2},
axis background/.style={fill=white}
]
\addplot [forget plot] graphics [xmin=-0.04, xmax=3.04, ymin=-4.05, ymax=3.05] {figures/walk_out1_lfm.pdf};
\end{axis}
\end{tikzpicture}%
 & % This file was created by matlab2tikz.
%
\begin{tikzpicture}
\tikzset{font=\small}
\begin{axis}[%
width=\figurewidth,
height=\figureheight,
at={(0\figurewidth,0\figureheight)},
scale only axis,
axis on top,
xmin=0,
xmax=3,
xtick={0,1,2,3},
xticklabels={},
ymin=-4,
ymax=3,
ytick style={draw=none},
ytick={-4,-3,-2,-1,0,1,2,3},
yticklabels={},
axis background/.style={fill=white}
]
\addplot [forget plot] graphics [xmin=-0.04, xmax=3.04, ymin=-4.05, ymax=3.05] {figures/walk_out1_kfflfm_S100.pdf};
\end{axis}
\end{tikzpicture}%
 & % This file was created by matlab2tikz.
%
\begin{tikzpicture}
\tikzset{font=\small}
\begin{axis}[%
width=\figurewidth,
height=\figureheight,
at={(0\figurewidth,0\figureheight)},
scale only axis,
axis on top,
xmin=0,
xmax=3,
xtick={0,1,2,3},
xticklabels={},
ymin=-4,
ymax=3,
ytick style={draw=none},
ytick={-4,-3,-2,-1,0,1,2,3},
yticklabels={},
axis background/.style={fill=white}
]
\addplot [forget plot] graphics [xmin=-0.04, xmax=3.04, ymin=-4.05, ymax=3.05] {figures/walk_out1_kfflfm_S10.pdf};
\end{axis}
\end{tikzpicture}%
\\
		\raisebox{.8cm}{\rotatebox{90}{lradius-Xrot}} & % This file was created by matlab2tikz.
%
\begin{tikzpicture}
\tikzset{font=\small}
\begin{axis}[%
width=\figurewidth,
height=\figureheight,
at={(0\figurewidth,0\figureheight)},
scale only axis,
axis on top,
xmin=0,
xmax=3,
xtick={0,1,2,3},
ymin=0,
ymax=100,
ytick style={draw=none},
ytick={0,20,40,60,80,100},
axis background/.style={fill=white}
]
\addplot [forget plot] graphics [xmin=-0.04, xmax=3.04, ymin=-0.05, ymax=100.05] {figures/walk_out2_lfm.pdf};
\end{axis}
\end{tikzpicture}%
 & % This file was created by matlab2tikz.
%
\begin{tikzpicture}
\tikzset{font=\small}
\begin{axis}[%
width=\figurewidth,
height=\figureheight,
at={(0\figurewidth,0\figureheight)},
scale only axis,
axis on top,
xmin=0,
xmax=3,
xtick={0,1,2,3},
ymin=0,
ymax=100,
ytick style={draw=none},
ytick={0,20,40,60,80,100},
yticklabels={},
axis background/.style={fill=white}
]
\addplot [forget plot] graphics [xmin=-0.04, xmax=3.04, ymin=-0.05, ymax=100.05] {figures/walk_out2_kfflfm_S100.pdf};
\end{axis}
\end{tikzpicture}%
 & % This file was created by matlab2tikz.
%
\begin{tikzpicture}
\tikzset{font=\small}
\begin{axis}[%
width=\figurewidth,
height=\figureheight,
at={(0\figurewidth,0\figureheight)},
scale only axis,
axis on top,
xmin=0,
xmax=3,
xtick={0,1,2,3},
ymin=0,
ymax=100,
ytick style={draw=none},
ytick={0,20,40,60,80,100},
yticklabels={},
axis background/.style={fill=white}
]
\addplot [forget plot] graphics [xmin=-0.04, xmax=3.04, ymin=-0.05, ymax=100.05] {figures/walk_out2_kfflfm_S10.pdf};
\end{axis}
\end{tikzpicture}%
\\  		
		& Time (s) & Time (s) & Time (s)
	\end{tabular}
	\caption{Comparison of the predictive GPs obtained for the the motion ``Walk'' using the standard LFM (first column) and the RFF approximation for $S=100$ (third column) and $S=10$ samples (third column) . Training data is represented using red dots and Test data using blue dots. The black line in the mean over the predictive GP function, and the shaded region denotes two times the standard deviation.}
	\label{fig:walk}
\end{figure*}
We remark that the evaluation of the covariance function ODE2 is the most expensive one because it requires the evaluation of the Faddeeva function. Hence, the computation time per iteration is reduced using the inner product of $v_d^{(2)}(t,\theta_d,\lambda)$.

\section{RANDOM FOURIER FEATURES FOR CONVOLVED MULTIPLE OUTPUT GAUSSIAN PROCESSES}
%\section{Random Fourier features for convolved multiple output GP}
Convolution processes can be used to build kernels for vector-valued functions, as reviewed in
\citet{Alvarez:CompEfficient:jmlr:2011}. Following similar expressions to the ones in section \ref{rfflfm}, an output
$f_d(\mathbf{x})$, with $\mathbf{x}\in\mathbb{R}^{p}$, can be modeled as a convolution integral of general smoothing
kernels $\{G_{d,q}^i(\cdot)\}_{d=1,q=1, i=1}^{D, Q, R_q}$, and latent processes $\{u_q^i(\mathbf{x})\}_{q=1, i=1}^{Q, R_q}$ 
\begin{align*}
f_d(\mathbf{x}) =
  \sum_{q=1}^Q\sum_{i=1}^{R_q}\int_{\mathcal{X}}G_{d,q}^i(\mathbf{x}-\mathbf{z})u_q^i(\mathbf{z})d\mathbf{z},
\end{align*}
where, according to \citet{Alvarez:CompEfficient:jmlr:2011}, the variable $R_q$ makes reference to the number of latent functions $u_q$ that share the same covariance function $k_q(x, x’)$, although are sampled independently. 
Granted that the $u_q^i(\mathbf{x})$ are independent GPs with zero mean and covariance functions
$\operatorname{cov}[u_q^i(\mathbf{x}), u_{q'}^{j}(\mathbf{x'})] = k_q(\mathbf{x},
\mathbf{x}')\delta_{q,q'}\delta_{i,j}$, where $\delta_{q,q'}$ and $\delta_{i,j}$ are Kronecker deltas, the cross-covariance between $f_d(\mathbf{x})$, and $f_{d'}(\mathbf{x}')$, $k_{f_d, f_{d'}}(\mathbf{x}, \mathbf{x}')$, follows a familiar form 
\[ \sum_{q=1}^Q\sum_{i=1}^{R_q}\int_{\mathcal{X}}G_{d,q}^i(\mathbf{x}-\mathbf{z})\int_{\mathcal{X}}G_{d',q}^i(\mathbf{x}'-\mathbf{z}')
 k_q(\mathbf{z}, \mathbf{z}')d\mathbf{z}d\mathbf{z}'.
\]
This covariance function subsumes several other covariance functions proposed in the literature for multiple output GPs, including the linear
model of coregionalization \citep{Alvarez:CompEfficient:jmlr:2011}.  

A general purpose expression for $k_{f_d, f_{d'}}(\mathbf{x}, \mathbf{x}')$ can be obtained by assuming that both
$G_{d,q}^i(\cdot)$ and $k_q(\cdot, \cdot)$ follow Gaussian forms. The cross-covariance $k_{f_d, f_{d'}}(\mathbf{x},
\mathbf{x}')$ would then also follow a Gaussian form after solving the double integration for
$\mathcal{X}=\mathbb{R}^p$. The authors in \citet{Alvarez:CompEfficient:jmlr:2011} provided a closed-form expression for
$k_{f_d, f_{d'}}(\mathbf{x}, \mathbf{x}')$ for this case, when $R_q=1$.

We can also use random Fourier features for $k_q(\cdot, \cdot)$ in the expression above. For the Gaussian case, since
the integrations are over $\mathbb{R}^p$, we use a Fourier transform instead of a Laplace transform as it was the case
for the LFM. Let us assume that both $G_{d,q}(\cdot)$ and $k_q(\cdot, \cdot)$ follow Gaussian forms,
\begin{align*}
G_{d,q}(\bm{\tau}) &= \exp\left(-\frac{P_d}{2}\bm{\tau}^\top\bm{\tau}\right),\\
  k_q(\boldz, \boldz') &= \exp\left(-\frac{1}{\ell^2_q}(\boldz-\boldz')^\top (\boldz-\boldz')\right),
\end{align*}
where $P_d$ is the inverse-width associated to the smoothing kernel for output $d$, and $\ell_q$ is the length-scale for
the kernel of the latent function.
The cross-covariance $k_{f_d, f_{d'}}(\mathbf{x}, \mathbf{x}')$ follows as
\begin{align*}
& \sum_{q =1}^Q S_{d,q}S_{d,q'}\int_{\mathcal{X}}                  \int_{\mathcal{X}}\exp\left(-\frac{P_d}{2}(\boldx-\boldz)^\top(\boldx-\boldz)\right)\\
 & \times \exp\left(-\frac{P_{d'}}{2}(\boldx'-\boldz')^\top(\boldx'-\boldz')\right)k_q(\boldz,\boldz')d\boldz d\boldz'.
\end{align*}
Using again the Bochner's theorem for $k_q(\boldz,\boldz')$,
\begin{align*}
  k_q(\boldz,\boldz') = \int p(\bm{\lambda}) \exp(j\bm{\lambda}^{\top}(\boldz-\boldz')) d \bm{\lambda}. 
 \end{align*} 
Placing this form for $k_q(\boldz,\boldz')$ inside the expression for $k_{f_d, f_{d'}}(\mathbf{x}, \mathbf{x}')$, and
solving the integral over $\bm{\lambda}$ using Monte Carlo, we get that  $k_{f_d, f_{d'}}(\boldx, \boldx')$ follows
\begin{align*}
 \sum_{q =1}^Q \frac{S_{d,q}S_{d,q'}}{S}\bm{\phi}^{\top}_d(\boldx, P_d, \bm{\Lambda}_q) \bm{\phi}^*_{d'}(\boldx', P_{d'}, \bm{\Lambda}_q), 
\end{align*}
where 
\begin{align*}
\bm{\phi}_d(\boldx, P_d, \bm{\Lambda}_q) = \exp\left(-\frac{1}{2P_d}\mathbf{b}_q + j\bm{\Lambda}_q\boldx\right),
\end{align*}
with $\mathbf{b}_q =  \sum_j \left(\bm{\Lambda}_q\odot\bm{\Lambda}_q\right)_{i,j}\in\mathbb{R}^{S\times 1}$, being $\odot$
the Hadamard product, and $\bm{\Lambda}_q = \frac{1}{\ell_q}\mathbf{Z}\in \mathbb{R}^{S\times p}$, where the entries of
the matrix $\mathbf{Z}$ are sampled from $\mathcal{N}(0,1)$. Hyperparameters $\theta_d$ and $\ell_q$ can be estimated using
similar procedures to the ones described in section \ref{sec:hyper}.

\paragraph{SARCOS} As an illustration of the use of the kernel above, we performed an experiment on a subset of the SARCOS dataset described in the book by \citet{Rasmussen:book06}.\footnote{Available at  \url{http://www.gaussianprocess.org/gpml/data/}} We use a subset of the data in the file \texttt{sarcos\_inv.mat}. In
particular, we randomly select $10000$ data observations that include two outputs, corresponding to the first two joint
torques, and the first seven inputs, corresponding to the joint positions. We then randomly select $1000$ observations
for the second output as the test data. We use the remaining $19000$ for training, this is, for hyperparameter
optimization. We compare the performance between the kernel proposed in \citet{Alvarez:CompEfficient:jmlr:2011} (CMOC)
and the kernel obtained using the random Fourier response features for different values of $S$. For the CMOC we optimize
the marginal likelihood as in Eq. \eqref{eq:full:marginal}, whereas for the RFRF, we use the marginal likelihood as in
Eq. \eqref{eq:low:rank:marginal}. 
Table \ref{tab:sarcos} reports the NMSE and NLPD for the $1000$ test observations for
the second output. These experiments were carried out using a single core of an Intel Xeon E5-2630v3 @ 2.4 GHz.

\begin{table}[h!]
	\caption{Results for the Sarcos Experiment.}
	\label{tab:sarcos}
	\setlength{\tabcolsep}{2pt}
	\centering
	\begin{tabular}{c|cc|c}
		\hline
		\hline		
	    Kernel & NMSE & NLPD  & Time [s]\\
		\hline	
        RFF+GG+S50   & 0.34 & 3.58 & 10.14 \\
		RFF+GG+S100  & 0.30 & 3.52 & 18.47 \\
		RFF+GG+S200  & 0.26 & 3.44 & 38.55 \\
        RFF+GG+S500  & 0.24 & 3.41 & 64.62\\
        RFF+GG+S1000 & 0.22 & 3.36 & 85.00\\
        CMOC         & \bfseries{0.19} & \bfseries{3.21} & 353.00 \\
		\hline
		\hline
	\end{tabular}
      \end{table}

      We notice that the performance of the approximation increases with $S$, and approaches the performance of CMOC,
      keeping the computation time per iteration to a fraction of the original one. As it was also expected, in higher
      dimensions, we need a larger number of random features to approach the performance of the CMOC.

\section{RELATED WORK}

Random Fourier features have been used in the literature for Gaussian processes before. For example, in \citet{Bonilla:2016}, the authors use RFFs in order to propose a multi-task GP model that circumvents the scalability problem of the GPs. Their model for the multiple outputs uses an affine transformation of the random features, whereas we use a non-instantaneous transformation via the Green's
functions. Also in \citet{Yang:2015}, the authors use a faster approximation of random Fourier features via the FastFood kernels \citep{Le:FastFood:2013}, for approximating the kernel functions of a GP. Their method is not used for multiple outputs, nor does include dynamical systems. 

Latent force models have been also studied using a state-space formulation \citep{Hartikainen:sequentialLFM:2011} and in
that line of research, low-rank approximations for computing features have also been introduced
\citep{Solin2014HilbertSM}. Specifically, this work approximates the covariance function using the Laplace operator eigenvalues and eigenfunctions. 
This formulation has been used in \citet{Svensson:2016} to approximate the GP priors that are
placed over the functions that transform the state vector in the update state and observation equations. Thus, it has not been considered to approximate the GP model of the excitation function.

\citet{Brault:2016} directly build random Fourier features for vector-valued kernels using an
operator-valued version of Bochner's theorem. The construction is applied to the decomposable kernel, the curl-free
kernel and the div-free kernel. In our construction, rather than starting with a fixed form for the
operator-valued kernel, we use a general mechanism used to build valid operator kernel functions and apply linear operators over
the random Fourier features defined for single output kernels. 
%\section{Experimental setup}
%\vspace{-.13cm}
\section{CONCLUSIONS AND FUTURE WORK}
%\section{Conclusions and Future Work}
We have shown in this paper how to use random Fourier features for easing the computation of the kernel functions
associated to LFMs. As a by-product, we have also reduced the computational complexity of working in
multiple-output GPs from $\mathcal{O}(D^3N^3)$ to $\mathcal{O}(DNQ^2S^2)$. We showed experiments over datasets of
different sizes for which results with LFM are slow to compute. Our random Fourier response features reduce
computational time without compromising performance. Also, notice that by having decoupled the solution of the
convolution integrals from the particular form for the kernel of the latent functions, we now can easily build kernels for
latent force models with different kernel functions in the GPs of the latent functions, just by changing the distribution
$p(\lambda)$ from which we sample from.

These novel representations of latent force models open the path for different types of future work: the application of
random Fourier response features for building more efficient versions of sequential LFM \citep{Alvarez:switched11} and
hierarchical LFM \citep{Honkela:PNAS10}; the use of physically inspired Fourier features in other Gaussian process
models, particularly, deep models \citep{Cutajar:rff:deepgps:2017}; the use of more efficient sampling techniques 
for obtaining the Fourier features, e.g. Quasi-Monte Carlo sampling \citep{Avron:QuasiMonteCarlo:FF:2016}. With a
more efficient way to compute kernels for multiple-outputs, we can also use more expensive model selection approaches,
for example, those based on automatic composition of kernel functions \citep{Duvenaud:AutomaticStats:2013}, for building
more complex covariance functions, e.g. combinations of first order models and second order models, as sums of kernels
or as products of kernels. For the case of convolved multiple outputs GPs where the input dimension is greater than
three (compared to typical LFMs), the computation of dense Gaussian matrices can be replaced by the product between
Hadamard matrices and diagonal Gaussian matrices, which are faster to compute \citep{Le:FastFood:2013}.

\subsubsection*{Acknowledgments} CG would like to thank to Convocatoria 567 from Administrative Department of Science,
Technology and Innovation of Colombia (COLCIENCIAS) for the support. MAA has been financed by the Engineering and Physical Research Council (EPSRC) Research Project EP/N014162/1.

\bibliography{bibkfflfms}
\bibliographystyle{plainnat}

\end{document}